\documentclass[letterpaper]{article} 
\usepackage{aaai2027}  
\usepackage[hyphens]{url}  
\usepackage{graphicx} 
\usepackage{natbib}  
\usepackage{caption} 
\usepackage{algorithm}
\usepackage{algorithmic}

\usepackage{newfloat}
\usepackage{listings}
\DeclareCaptionStyle{ruled}{labelfont=normalfont,labelsep=colon,strut=off} 
\floatstyle{ruled}
\newfloat{listing}{tb}{lst}{}
\floatname{listing}{Listing}

\usepackage{booktabs}

\usepackage{bm}
\usepackage{amssymb}
\usepackage{amsmath}
\usepackage{comment}
\usepackage{tabularx}
\usepackage{bbding}
\usepackage{multirow}

\title{Visual Representation Matters: Exploiting Temporal Differences in Video-to-Audio Generation}
\author{
    Zehua Chen\textsuperscript{\rm 1}\thanks{Equal contribution. E-mails: zhc23thuml@tsinghua.edu.cn, ackokomi0222@gmail.com.}, Junyou Wang\textsuperscript{\rm 1}\footnotemark[1], Yuxuan Jiang\textsuperscript{\rm 1}, Zhenying Fang\textsuperscript{\rm 2},\\
    Yusheng Dai\textsuperscript{\rm 3}, Jianfei Chen\textsuperscript{\rm 1}, Ziwei Liu\textsuperscript{\rm 4}, Jun Zhu\textsuperscript{\rm 1}\thanks{Corresponding author. E-mail: dcszj@tsinghua.edu.cn.}
}
\affiliations{
    \textsuperscript{\rm 1}Tsinghua University, Beijing, China \quad
    \textsuperscript{\rm 2}Hefei University of Technology, Hefei, China\\
    \textsuperscript{\rm 3}Monash University, Australia \quad
    \textsuperscript{\rm 4}Nanyang Technological University, Singapore
}

\begin{document}

\maketitle

\begin{abstract}
Video-to-audio (V2A) generation extends image-to-audio generation (I2A) by introducing consecutive frames that provide essential temporal cues for audio synthesis. 
However, existing conditional diffusion-based V2A methods typically enhance visual conditioning with additional audio-visual supervision, acoustic structure prediction, or reasoning from large multimodal models, requiring extra networks or strong inductive biases. 
Inspired by recent advances in visual representation learning, we introduce TD-V2A, which leverages temporal differences (TD) as the key representation that distinguishes V2A from I2A, enriching visual conditioning with minimal architectural modification. 
We first investigate TD at both the frame and feature levels to identify the most effective representation level at which TD complements visual representations.
Based on these findings, we develop a hierarchically continual learning strategy and an annealed temporal differences guidance method to progressively learn and exploit TD information during diffusion training and sampling process, respectively. 
Extensive experiments on benchmark datasets demonstrate that effectively exploiting TD through our proposed framework significantly improves end-to-end V2A generation quality, even outperforming dedicated V2A representations such as contrastive audio-visual pretraining.
\end{abstract}


\section{Introduction}
\label{sec:intro}

Generating audio from visual information has attracted increasing attention for its potential in augmented reality, film production, and multimodal content creation~\cite{soundofpixels,luo2023diff,omni2sound}. Early studies primarily focused on image-to-audio (I2A) generation~\cite{TamingI2A,AudioCLIP,sheffer2023hear}, where audio is synthesized from a single static image. 
More recently, research has progressed toward video-to-audio (V2A) generation, which aims to generate realistic and coherent audio for dynamic video inputs. Unlike I2A, V2A requires not only semantic consistency between visual content and generated audio but also temporal alignment between evolving visual events and corresponding sounds~\cite{luo2023diff,wang2024tiva,cheng2025mmaudio,tian2025audiox,thinksound}. 
This additional challenge stems from the temporal differences (TD) between consecutive video frames, which characterize scene evolution over time. 
Without TD, a video reduces to a static image, and V2A inevitably collapses to an I2A setting~\cite{framebridge,daithankar2026you}.

Modern V2A systems are predominantly built upon conditional latent diffusion models, where the quality of the conditioning largely determines the fidelity of generated audio. 
Existing methods therefore focus on enriching the conditioning information through three main strategies: introducing auxiliary audio-oriented supervision, such as coarse mel-spectrograms or temporal audio structures~\cite{wang2024tiva,jeong2024rewas,comunita2024syncfusion,ren2025stav2a}; incorporating separately pre-trained audio-visual representation models~\cite{luo2023diff,iashin2024synchformer}; or leveraging multi-modal large language models for high-level video understanding~\cite{thinksound,omni2sound}. Although effective, these approaches rely on additional supervision, specialized pre-trained models, or strong inductive biases beyond the input video itself. 
In this work, inspired by recent advances in TD-based visual representation learning~\cite{daithankar2026you}, we instead exploit the inherent TD between video frames to enhance visual representations. As the key factor distinguishing V2A from I2A, TD provides an intrinsic cue to enable superior V2A quality without introducing additional predictive networks or external conditioning signals.

\begin{figure*}[h]
  \centering
  \includegraphics[width=1.0\linewidth]{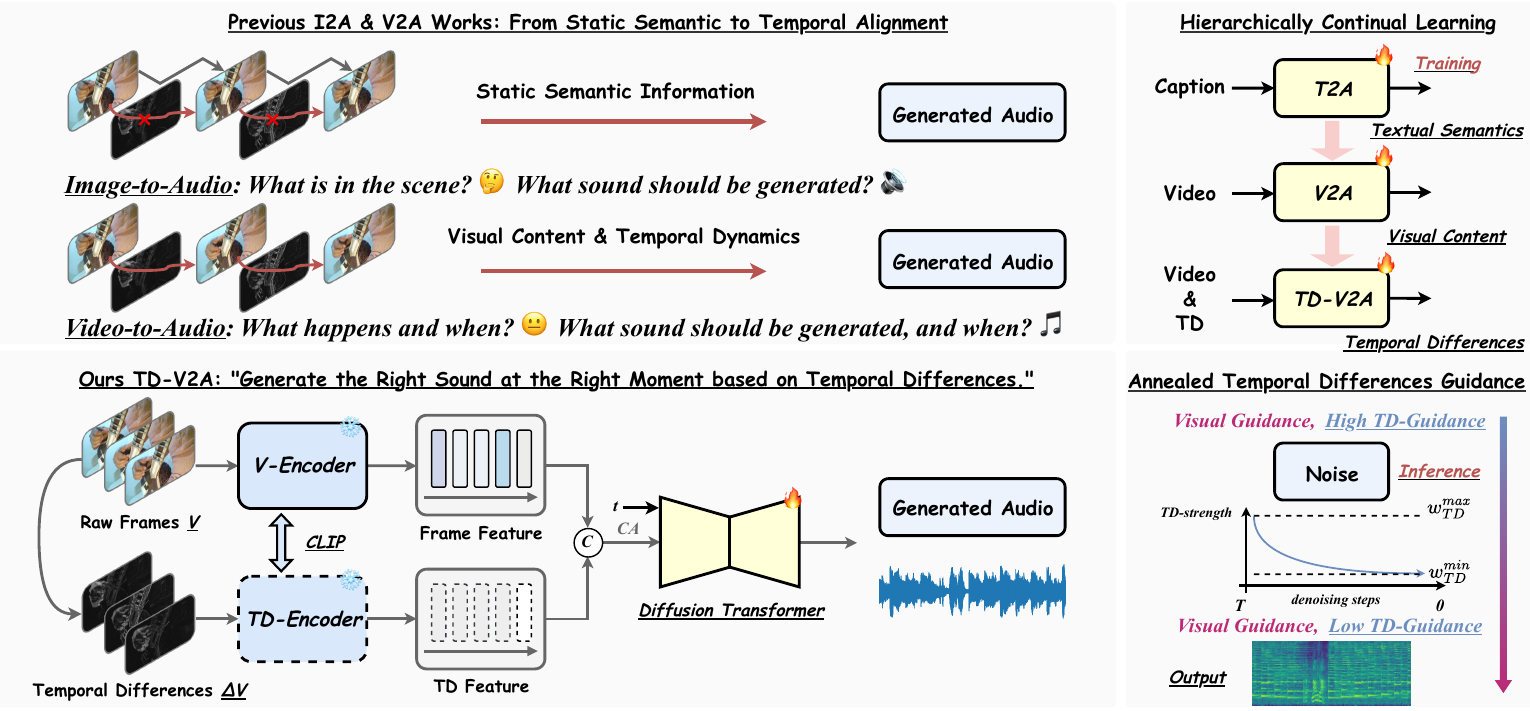}
\caption{Overview of TD-V2A. Left: The framework exploits temporal differences (TD) for enriching visual representation to enhance video-to-audio generation. Right: A hierarchically continual learning strategy progressively learns textual semantics, visual content and TD, while annealed temporal differences guidance further exploits TD by dynamically adjusting its guidance strength across sampling steps.}
\label{fig:main}
\end{figure*}

Motivated by this, we propose TD-V2A, which directly exploits TD representations derived from the input video without introducing additional supervision or dedicated extraction networks. 
Since TD can be computed in diverse representation spaces, we begin with an investigation of its form at two levels: frame-level TD and feature-level TD. 
Frame-level TD is obtained by differencing consecutive video frames before visual encoding, preserving complete visual variation while retaining the original visual input structure. This makes TD naturally compatible with pretrained visual encoders, allowing temporal cues to be incorporated with minimal architectural modification and inductive bias.
In contrast, feature-level TD is computed from the differences between encoded frame representations, where temporal variation is measured in a compressed semantic space. 
Although such representations capture high-level semantic transitions, the encoding process inevitably compresses and abstracts the original visual information, potentially discarding fine-grained temporal variations. 
Our investigation demonstrates that preserving TD before visual encoding provides substantially more informative conditioning for V2A generation than modeling TD in the compressed feature space.

When exploiting TD, the hierarchical structure of visual representations naturally motivates a progressive learning paradigm, where TD representations are progressively incorporated~\cite{ControlAudio,omni2sound}. 
Therefore, at the model training stage, we develop a Hierarchically Continual Learning (HCL) strategy for the diffusion models developed in a latent space compressed from waveform.
Specifically, HCL gradually exploits increasingly informative visual representations through three training stages. 
The first stage initializes the model with large-scale text-audio pre-training to acquire high-quality audio priors that facilitate subsequent V2A learning. 
The second stage adapts the model to V2A generation using raw visual representations. 
Finally, the third stage jointly optimizes the model under both raw visual conditioning and TD-augmented conditioning, enabling the conditional diffusion model to explicitly distinguish and effectively exploit TD for V2A generation.

At inference time, the hierarchical visual conditioning introduced by TD further inspires an Annealed Temporal Differences Guidance (ATDG) strategy. Combined with the coarse-to-fine denoising trajectory of diffusion models, ATDG progressively adjusts the influence of TD throughout the sampling process. Specifically, stronger TD guidance is applied during the early sampling stages to encourage the establishment of globally consistent temporal dynamics. As sampling proceeds toward finer details, the guidance strength is gradually annealed, allowing the generation process to rely increasingly on raw visual conditioning. By aligning the contribution of TD with the intrinsic denoising trajectory, ATDG enables the model to exploit TD information more effectively, resulting in more temporally coherent V2A generation.

Following previous works, we develop TD-V2A on the VGGSound and AudioSet benchmarks. Without additional supervision or strong inductive biases, we introduce TD as an effective means to enhance visual representations for end-to-end V2A generation. Equipped with the proposed HCL and ATDG strategies, TD-V2A achieves superior results with a commonly used DiT backbone~\cite{evans2024stableaudio,omni2sound}, validated by both objective and subjective evaluations. Notably, it even outperforms the additionally developed CAVP model~\cite{luo2023diff}, highlighting TD as a simple yet powerful alternative to strengthen V2A quality.

\section{Related Work}
\label{sec:relatedwork}

\paragraph{Temporal Differences.}
TD modeling is a fundamental learning principle that exploits temporal consistency to learn predictive representations from sequential data. Originally introduced in reinforcement learning through bootstrapping~\cite{sutton1988learning}, TD enables models to refine current estimates using future predictions without requiring complete future supervision. 
Owing to its generality, the TD principle has been successfully extended beyond reinforcement learning to generative modeling~\cite{jesse2025temporal} and visual representation learning,~\textit{i.e.}, TDV~\cite{daithankar2026you}, where temporal transitions provide intrinsic supervisory signals for learning meaningful features. 
These developments demonstrate that TDs encode rich structural information about the underlying dynamics, making TD a powerful framework for exploiting sequential observations. 
Inspired by this principle, our TD-V2A exploits temporal variations between consecutive video frames as informative cues to enhance visual representation and consequently enhance V2A generation results.

\noindent
\paragraph{Visual Representation in V2A Generation.}
Existing V2A generation methods typically improve generation quality by introducing additional supervision or strong inductive biases beyond raw video inputs, such as separately trained audio-visual representation models~\cite{luo2023diff, iashin2024synchformer}, auxiliary networks that predict intermediate audio structures (e.g., coarse mel-spectrograms~\cite{wang2024tiva} or temporal energy patterns~\cite{jeong2024rewas, comunita2024syncfusion, ren2025stav2a}), or large language models for high-level video understanding~\cite{thinksound,omni2sound}. 
While effective, these approaches require specialized models or handcrafted conditioning signals. 
In contrast, our method directly exploits natural temporal differences inherent in the input video, using frame-level temporal variations as intrinsic cues to enhance video understanding without introducing additional predictive networks or external conditions, thereby reducing reliance on strong inductive biases while achieving superior V2A generation.

\section{Preliminary}
\label{sec:preliminary}
    
\paragraph{Video-to-Audio Generation.}
\label{sec:v2a}
V2A generation aims to synthesize audio $\bm{x}_0$ that is both semantically and temporally aligned with a given silent video input $\bm{v}$~\cite{luo2023diff}.
Recent V2A methods~\cite{luo2023diff, jeong2024rewas, cheng2025mmaudio, tian2025audiox, omni2sound} typically adopt a conditional diffusion framework~\cite{liu2023audioldm,evans2024stableaudio}, where audio is generated by progressively denoising a latent representation under the guidance of a video condition $\bm{c}_{\text{video}}(\bm{v})$ extracted from the input video. 
During training, the model learns the correspondence between video and audio, and during inference, the extracted video condition controls the generation process toward aligned audio outputs. 
Formally, the diffusion training objective is defined as follows.
\begin{equation}
\mathcal{L}_{\text{v2a}} = \mathbb{E}_{\bm{z}_t, \bm{\epsilon}, \bm{c}_{\text{video}}} \Big[ \big\| \bm{\epsilon} - \bm{\epsilon}_\theta( \bm{z}_t, t, \bm{c}_{\text{video}}(\bm{v})) \big\|_2^2 \Big],
\end{equation}
where $\bm{z}_t$ is the noisy representation at time step $t$ constructed with the diffusion forward process in a small latent space $\bm z$ compressed from the audio space $\bm{x}$.
As shown, given this formulation, the quality of the generated audio naturally depends on how effectively the video condition captures the underlying visual content and temporal dynamics. Therefore, improving video representation is a critical factor for achieving high-quality V2A generation.

\paragraph{Additional Supervision.}
\label{sec:v2a}
Existing V2A methods commonly enhance generation quality by introducing auxiliary supervision beyond the original video condition. Representative approaches employ separately trained audio-visual representation models for enriched cross-modal conditioning, predict intermediate audio representations (e.g., mel-spectrograms or temporal energy patterns) with auxiliary networks, or leverage large language models for high-level video understanding and reasoning. Despite their different implementations, these methods share a common principle: they introduce additional modules to construct a strong auxiliary condition $\bm{c}_{\text{aux}}(\bm{v})$, which complements the original video representation during audio generation. The resulting diffusion objective can be generally formulated as follows.
\begin{equation}
\mathcal{L}_{\text{aux-v2a}} = \mathbb{E}_{\bm{z}_t, \bm{\epsilon}, \bm{c}_{\text{video}}, \bm{c}_{\text{aux}}} \Big[ \big\| \bm{\epsilon} - \bm{\epsilon}_\theta( \bm{z}_t, t, \bm{c}_{\text{video}}(\bm{v}), \bm{c}_{\text{aux}}(\bm{v})) \big\|_2^2 \Big].
\end{equation}
While effective, these approaches rely on additional predictive models, handcrafted supervision, or external reasoning modules, leading to increased engineering complexity and stronger inductive biases. 
In contrast, we improve the quality of the original video representation itself through enhanced video representation learning. By strengthening visual representation directly from the input video, our method improves V2A without introducing additional auxiliary networks.

\section{Method}
\label{sec:method}

\subsection{Motivation: TD Distinguish V2A from I2A}
Existing V2A methods primarily improve video-audio correspondence by introducing additional supervision or auxiliary modules with stronger inductive biases. 
In contrast, our goal is to enhance V2A generation from a more fundamental perspective: strengthening the video representation itself. 
We argue that the key lies in explicitly modeling TD, which constitute the essential distinction between V2A and I2A generation.~\textit{If a video contains no TD across frames, it effectively degenerates into a static image~\cite{framebridge,daithankar2026you}, reducing V2A to an I2A problem where temporal alignment is no longer a meaningful objective.} 
This observation suggests that TD are not merely complementary cues, but the intrinsic visual information capturing temporal dynamics. 
Motivated by this insight, we explicitly exploit TD from the input video to enable stronger visual representations with weaker inductive bias. 
By improving the quality of the original video representation rather than introducing additional auxiliary conditions, we achieve more temporally coherent and semantically faithful V2A generation in a simple end-to-end framework.

\subsection{Investigation of TD Representation}
\label{sec:first-order}

\begin{figure}[t]
  \centering
  \includegraphics[width=1.0\linewidth]{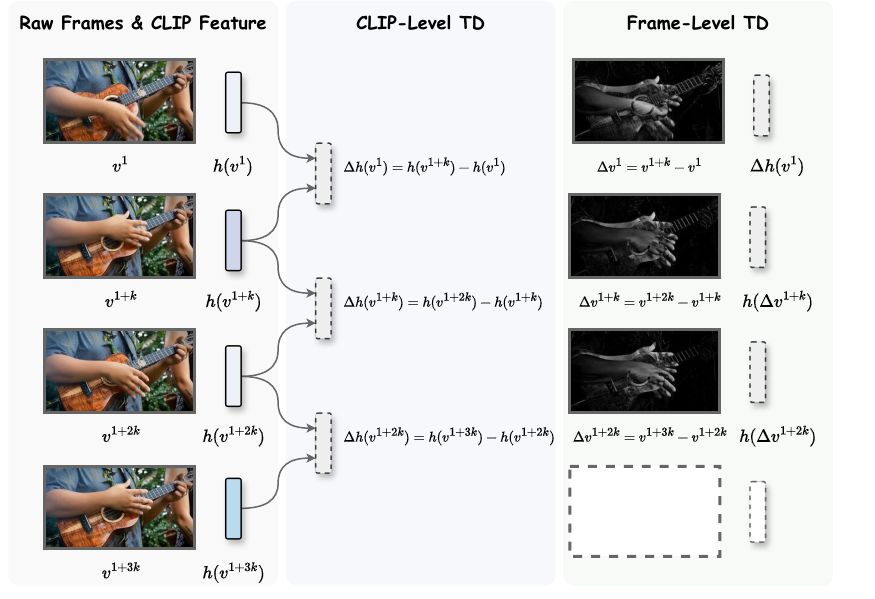}
\caption{Illustration of TD modeling at different granularities. While CLIP-level TD (middle) captures high-level semantic variations between embeddings, frame-level TD (right) explicitly characterizes temporal dynamics by computing pixel-wise residuals between raw frames.}
\label{fig:varyrepresentation}
\end{figure}

We begin by exploring TD representations at different levels of visual abstraction. Specifically, TD can be constructed either directly from consecutive video frames (frame-level TD) or from their corresponding visual features (feature-level TD). 
Since CLIP embeddings are the de facto visual representation in modern V2A systems~\cite{wang2024tiva,cheng2025mmaudio,tian2025audiox,omni2sound}, we instantiate feature-level TD as differences between consecutive CLIP embeddings, introducing temporal information with minimal architectural modification and without imposing additional inductive biases. In the following, we present the formulations of frame-level TD (FTD) and CLIP-level TD (CTD), respectively.

\subsubsection{Frame-Level Temporal Differences (FTD).}

At the frame level, TD can be obtained by computing differences between consecutive raw frames. 
Formally, given a sequence of video frames $\{\bm{v}^{n}\}_{n=1}^{N}$, the FTD computation is defined as:
\begin{equation}
\begin{aligned}
\label{framefo}
\Delta \bm{v}^{n} = \bm{v}^{n+k} - \bm{v}^{n},
\quad
\bm{c}_{\text{video}}^{\text{TD}}(\bm v) = [\bm{h}(\bm v^{n});\bm{h}(\Delta \bm v^{n})],
\end{aligned}
\end{equation}
where $ n=1, 2, \dots, N-k$; $k$ is a hyperparameter with a small value controlling the window length when computing frame differences; and $\bm{c}_{\text{video}}^{\text{TD}}$ is the computed video condition that incorporates TD.
Considering each element in $\Delta \bm{v}^{n}$, its amplitude reflects the magnitude of local temporal variations between consecutive frames. Different from raw frames that are dominated by static appearance information, FTD naturally suppresses temporally invariant components and highlights dynamic changes~\cite{daithankar2026you}. 
Since adjacent video frames usually exhibit strong temporal consistency, the resulting frame differences form a compact representation of visual evolution, where redundant background information is largely removed while informative motion patterns are preserved. Therefore, FTD provides a complementary visual representation that augments appearance features with explicit temporal variation cues, enabling V2A models to better capture the dynamic characteristics inherent in video inputs.

To enrich visual representations with explicit temporal variations without imposing strong inductive biases, we directly encode both raw frames $\bm{v}^{n}$ and and their temporal differences $\Delta \bm{v}^{n}$ using a shared pretrained CLIP encoder $\bm h$.  
Since temporal differences remain visual observations that describe the evolution of scene content, they can be naturally embedded into the same representation space as raw frames, eliminating the need for dedicated motion encoders or additional architectural modifications.
Following previous works~\cite{tian2025audiox, omni2sound}, the CLIP encoder is kept frozen during diffusion model training, allowing TD to be seamlessly incorporated into existing V2A frameworks while preserving their original visual backbone.

\subsubsection{CLIP-Level Temporal Differences (CTD).}

Beyond differences between raw frames, temporal variation can also be characterized directly in the CLIP embedding space.
Specifically, given the CLIP representations $\{\bm{h}(\bm{v}^{n})\}_{n=1}^{N}$ of video frames $\{\bm{v}^{n}\}_{n=1}^{N}$, we compute their temporal differences at the CLIP level as:
\begin{equation}
\label{clipfo}
\Delta \bm{h}(\bm v^{n}) = \bm{h}(\bm v^{n+k}) - \bm{h}(\bm v^{n}),
\bm{c}_{\text{video}}^{\text{TD}} (\bm v) = [\bm{h}(\bm v^{n});\Delta \bm{h}(\bm v^{n})],
\end{equation}
where $n=1,2, \dots ,N-k$; $k$ again denotes the temporal window length for computing the difference in the CLIP embedding space. 
Compared to FTD $\Delta \bm{v}^{n}$, CTD $\Delta \bm{h}(\bm v^{n})$ are computed after visual information has been projected into a high-level semantic representation. Consequently, they primarily capture semantic evolution while discarding a substantial amount of low-level visual variation. Although this abstraction improves semantic consistency, it may also attenuate fine-grained temporal changes that are preserved in FTD. Therefore, CTD provide a complementary view of temporal variation, emphasizing semantic transitions rather than detailed visual dynamics.

\subsubsection{TD Encoder.}
We encode both raw frames and TD using the same pretrained CLIP encoder, treating TD as a complementary visual representation rather than introducing additional TD-specific encoders. 
Although it is possible to develop dedicated encoders for TD or motion representations, doing so inevitably introduces additional architectural complexity and stronger task-specific inductive biases.
This design is supported by CLIP's strong visual generalization capability~\cite{clipicml}. For example, CLIP achieves robust recognition on ImageNet-Sketch~\cite{wang2019learning}, indicating that its representations remain effective even when visual appearance is largely reduced to contours and structural cues~\cite{clipicml}. Since FTD similarly suppress static appearance while emphasizing structural changes, they can be naturally encoded within the same visual representation space. We therefore retain the standard frozen CLIP encoder throughout, enabling a controlled evaluation of TD under minimal architectural modification and inductive bias.

\subsection{Hierarchically Continual Learning}
\label{sec:integration}

Raw frames and TD provide hierarchical visual representations from appearance to temporal variation, naturally motivating a progressive learning strategy to fully exploit visual information. Based on this observation, we develop a three-stage hierarchical training framework that incrementally incorporates more detailed visual representations.

\subsubsection{Textual Semantics.}
\label{sec:t2a}
The T2A pre-training stage provides a strong initialization for V2A, allowing the model to acquire advanced audio generation capabilities before introducing visual conditioning. In this stage, we adopt Stable Audio~\cite{evans2024fast} as the T2A backbone, which employs a DiT-based latent diffusion architecture operating on a compressed audio representation derived directly from waveform signals. Its powerful audio modeling capability provides a strong foundation for subsequent video-conditioned audio generation.
Formally, the T2A LDM is pretrained to generate a latent audio representation $\bm{z}_0$ conditioned on a text embedding $\bm{c}_\text{text}$, with the objective:
\begin{equation}
\mathcal{L}_{\text{Stage1-T2A}} = \mathbb{E}_{\bm{z}_0, \bm{\epsilon}, t} \Big[ \big\| \bm{\epsilon} - \bm{\epsilon}_\theta(\bm{z}_t, t, \bm{c}_\text{text}) \big\|_2^2 \Big],
\end{equation}
where $\bm{c}_\text{text}$ can be extracted with either contrastive pre-training models,~\textit{e.g.}, CLAP~\cite{wu2023clap}, or language models,~\textit{e.g.}, FLAN-T5~\cite{chung2024scaling}. 

\subsubsection{Visual Content.}
\label{sec:v2a-ft}

Starting from the pretrained T2A model, we further adapt it to V2A generation by replacing the text condition with visual representations. Since T2A and V2A share the same diffusion framework~\cite{tian2025audiox,omni2sound}, the pretrained audio generation capability can be naturally transferred to video-conditioned generation. 
Formally, the fine-tuning objective replaces the text embedding $\bm{c}_\text{text}$ with a visual embedding,~\textit{e.g.}, CLIP embeddings~\cite{clipicml}, extracted from raw video frames:
\begin{equation}
\mathcal{L}_{\text{Stage2-V2A}} = \mathbb{E}_{\bm{z}_0, \bm{\epsilon}, t} \Big[ \big\| \bm{\epsilon} - \bm{\epsilon}_\theta(\bm{z}_t, t, \bm{c}_{\text{video}}) \big\|_2^2 \Big].
\end{equation}

Following recent V2A and VT2A generation systems~\cite{wang2024tiva,cheng2025mmaudio,tian2025audiox,omni2sound}, we use CLIP~\cite{clipicml} to extract visual embeddings from raw video frames as the conditioning signal. The visual-conditioned fine-tuning enables the model to establish video-audio correspondence while retaining the audio generation prior learned during T2A pre-training.

\subsubsection{Temporal Differences.}
\label{sec:v2a-ft}

After obtaining the visual-conditioned V2A model, we further introduce TD to enrich the visual representation. Rather than replacing raw frame features, TD serves as a complementary representation that explicitly characterizes frame-to-frame changes while remaining derived from the same visual observations. 
Since the model has already learned the basic visual representation through V2A fine-tuning, TD is introduced only in the third stage. Specifically, we jointly optimize both the original V2A task and the TD-enhanced V2A task, allowing the model to progressively learn the contribution of TD while preserving its original visual representation capability. The TD-enhanced objective is formulated as follows.
\begin{equation}
\mathcal{L}_{\text{Stage3-TDV2A}} = \mathbb{E}_{\bm{z}_0, \bm{\epsilon}, t} \Big[ \big\| \bm{\epsilon} - \bm{\epsilon}_\theta(\bm{z}_t, t, \bm{c}_\text{video}^{\text{TD}}) \big\|_2^2 \Big].
\end{equation}
where $\bm{c}_\text{video}^{\text{TD}}$ denotes the video condition that integrates TD at frame or CLIP level, computed as Equation~\eqref{framefo} or Equation~\eqref{clipfo} defined above. 
\begin{table*}[t]
\centering
\small
\begin{tabular}{l | c c c | c c c c | c | c}
\toprule
\multirow{2}{*}{\textbf{Method}} & \multirow{2}{*}{\textbf{Pre-training}} & \multirow{2}{*}{\textbf{Condition}} & \multirow{2}{*}{\textbf{Guidance}} & \multicolumn{4}{c}{\textbf{Audio Quality}} & \multicolumn{1}{c}{\textbf{Semantic}} & \multicolumn{1}{c}{\textbf{Temporal}} \\
\cmidrule(lr){5-8} \cmidrule(lr){9-9} \cmidrule(lr){10-10}
& & & & \textbf{FAD $\downarrow$} & \textbf{KL $\downarrow$} & \textbf{IS $\uparrow$} & \textbf{FD $\downarrow$} & \textbf{IBS $\uparrow$} & \textbf{AA $\uparrow$} \\
\midrule
GT & - & - & - & - & - & - & - & 32.9 & 83.6 \\
\midrule
IM2WAV & \XSolidBrush & CLIP & CFG & 6.41 & 2.54 & - & - & 19.0 & 74.3  \\
Diff-Foley & T2I & CAVP$^\dagger$ & CFG+CG & 5.79 & 3.12 & 10.8 & 21.90 & 20.4 & \textbf{89.9} \\
FoleyGen & \XSolidBrush & CLIP & CFG & 1.65 & 2.35 & - & - & 26.1 & 73.8 \\
VTA-LDM & Unlabeled V/A & CLIP4CLIP & CFG & 2.01 & 2.37 & 10.4 & 12.80 & 26.2 & 77.0 \\
FoleyCrafter & T2A & CLIP+Timestamp$^\dagger$ & \XSolidBrush & 2.32 & 2.54 & 9.9 & 18.10 & 27.7 & 83.6 \\
Frieren & \XSolidBrush & CAVP & CFG & 1.23 & 2.73 & 11.3 & 12.13 & 21.0 & - \\
V2A-Mapper & \XSolidBrush$^*$ & CLIP (CLAP)$^{\dagger**}$ & CFG & 0.90 & 2.68 & \underline{12.5} & 8.35 & 22.4 & 78.3 \\
VAB-Encodec & V-A pairs & eva-CLIP & CFG & 2.69 & 2.58 & - & - & - & - \\
VATT$^{\diamond}$ & V-A pairs & Trained visual feature$^\dagger$ & CFG & 2.35 & 2.25 & - & - & - & 82.8  \\
MMAudio$^{\diamond}$ & \XSolidBrush & CLIP+Syncformer$^\dagger$ & CFG & \underline{0.81} & \textbf{2.11} & 11.9 & \underline{5.65} & \underline{28.0} & - \\
AudioX$^{\diamond}$ & T2A & CLIP+Syncformer$^\dagger$ & CFG & 1.13 & 2.57 & 12.2 & 8.83 & 26.0 & - \\
\midrule
TD-V2A (Ours) & HCL & CLIP+FTD & ATDG & \textbf{0.53} & \underline{2.16} & \textbf{16.9} & \textbf{3.79} & \textbf{33.8} & \underline{89.1} \\
\bottomrule
\end{tabular}
\caption{Comparison of our method with baseline models on the VGGSound test set. The best performance for each metric is in bold, while the second-best is marked with an underline. In the "Pre-training" column, "T", "I", "A", "V" stands for text, image, audio and video, respectively.$^{\diamond}$: does not use text during inference. $^{\dagger}$: the feature requires additional network or training. $^{*}$: V2A-Mapper directly uses T2A (AudioLDM) for generation. $^{**}$: The mapper translating CLIP into CLAP needs training.
}
\label{tab:v2aobjective}
\end{table*}

\subsection{Annealed Temporal Differences Guidance}
\label{sec:inference}
During inference, diffusion models progressively reconstruct structured signals from Gaussian noise through a coarse-to-fine denoising process. In our framework, TD provide a complementary visual representation that captures frame-to-frame variations beyond the semantic information encoded by raw frame features. While HCL enhances the learning of TD representations during training, we further exploit these representations during inference through an adaptive guidance strategy.
Specifically, the jointly optimized model provides two complementary conditional predictions: one conditioned on raw frame representations and the other on TD-enhanced representations. The difference between them reflects the contribution of TD representations, which serves as an explicit temporal guidance signal during sampling.
Let $\bm{\epsilon}_{\bm{\theta}}(\bm{z}_t, t)$ denote the unconditional prediction, $\bm{\epsilon}_{\bm{\theta}}(\bm{z}_t, t, \bm{c}_{\text{video}})$ the prediction conditioned on frame-level features, and $\bm{\epsilon}_{\bm{\theta}}(\bm{z}_t, t, \bm{c}_{\text{video}}^{\text{TD}})$ the prediction conditioned on both raw frame and temporal differences. The final guided prediction is computed as follows.
\begin{equation}
\begin{aligned}
\bm{\epsilon}_{\bm{\theta}}^{\text{ATDG}} 
&= \bm{\epsilon}_{\bm{\theta}}(\bm{z}_t, t)+
w_f \Big( \bm{\epsilon}_{\bm{\theta}}(\bm{z}_t, t, \bm{c}_{\text{video}}^{\text{TD}}) - \bm{\epsilon}_{\bm{\theta}}(\bm{z}_t, t) \Big) \\
&+w_{\text{TD}}(t) \cdot \Big( \bm{\epsilon}_{\bm{\theta}}(\bm{z}_t, t, \bm{c}_{\text{video}}^{\text{TD}}) - \bm{\epsilon}_{\bm{\theta}}(\bm{z}_t, t, \bm{c}_{\text{video}}) \Big),
\end{aligned}
\end{equation}
where $w_f$ is a constant guidance scale for frame-level semantics, and $w_{\text{TD}}(t)$ is a timestep-dependent weight for TD.
To emphasize temporal guidance at early stages, we define:
\begin{equation}
w_{\text{TD}}(t) = w^{\text{min}}_{\text{TD}} + (w^{\text{max}}_{\text{TD}} - w^{\text{min}}_{\text{TD}}) \cdot (\frac{t}{T})^{\gamma},
\end{equation}
where $T$ is the total number of sampling steps, $t$ decreases from $T$ to $0$, and $\gamma$ is the annealing factor.
Under this formulation, TD dominate early steps and are gradually annealed, allowing semantic guidance to take over in later stages. 
This design follows the coarse-to-fine nature of diffusion sampling, improving temporal alignment while preserving semantic fidelity. It can be viewed as a mixture-of-guidance strategy~\cite{AudioMoG}, where variation and semantic cues contribute complementarily, yielding targeted alignment gains without additional inference cost, as demonstrated by our experimental results.

\section{Experiment}
\label{sec:experiment}

\subsection{Experimental Setup}
\label{sec:setup}

\subsubsection{Datasets.} For T2A pre-training, we utilize AudioCaps training set \cite{kim2019audiocaps}, AudioSet \cite{gemmeke2017audio}, VGGSound training set \cite{chen2020vggsound}, FreeSound \cite{font2013freesound}, and MSD \cite{bertin2011million}. For visual and temporal differences fine-tuning, we use AudioSet and VGGSound training sets (detailed in Appendix~\ref{app:datasets}). All audio tracks are segmented into 10-second clips and resampled to 16 kHz. 
Following previous works~\cite{luo2023diff,tian2025audiox,zhang2024foleycrafter}, we evaluate on the VGGSound test set, consisting of about 15K 10-second audio clips. 

\begin{table}[t]
\centering
\small
\setlength{\tabcolsep}{4pt}
\begin{tabular}{l | c c c}
\toprule
\textbf{Method} & \textbf{OVL $\uparrow$} & \textbf{S-REL $\uparrow$} & \textbf{T-REL $\uparrow$}\\
\midrule
GT & $3.81 \pm 0.49$ & $3.92 \pm 0.55$ & $3.72 \pm 0.47$ \\ 
\midrule
Diff-Foley & $2.46 \pm 0.49$ & $2.59 \pm 0.54$ & $2.21 \pm 0.45$ \\
FoleyCrafter & $3.17 \pm 0.47$ & $3.23 \pm 0.56$ & $3.04 \pm 0.50$ \\
\midrule
TD-V2A (Ours) & $\bm{3.68 \pm 0.46}$ & $\bm{3.85 \pm 0.51}$ & $\bm{3.62 \pm 0.48}$ \\
\bottomrule
\end{tabular}
\caption{Subjective evaluation results for different methods.}
\label{tab:subjective}
\end{table}

\subsubsection{Model Training.}
During T2A pre-training, we use FLAN-T5~\cite{chung2024scaling} as the text encoder, as it provides a stronger initialization for subsequent fine-tuning than CLIP or CLAP. We first train a Variational Autoencoder (VAE) to compress waveforms into latent representations, and pretrain the T2A model for 2M iterations with a total batch size of 64. For visual and temporal differences fine-tuning, we employ the same VAE and Diffusion Transformer (DiT) \cite{peebles2023scalable} backbone as in T2A, and each for 0.3M iterations with the same total batch size on 8 GPUs. We use the AdamW optimizer with a learning rate of $5\times 10^{-5}$. Implementation details are further provided in Appendix~\ref{app:vae} and Appendix~\ref{app:ldm}.

\subsubsection{Model Inference.}
To apply ATDG, we train with a dropout probability of 0.1 on the CLIP condition $\bm{c}_\text{video}$ during visual fine-tuning. In the final fine-tuning with temporal differences, we apply 0.05 dropout to the TD condition and another 0.05 dropout to both conditions. The reason for using smaller probabilities is that $\bm{\epsilon}_{\bm{\theta}}(\bm{z}_t, t)$ and $\bm{\epsilon}_{\bm{\theta}}(\bm{z}_t, t, \bm{c}_{\text{video}})$ have already been trained in the previous visual fine-tuning stage.
We set $w_f=2.0, w^{\text{min}}_{\text{TD}}=0.5, w^{\text{max}}_{\text{TD}}=1.5$ and $\gamma=1.2$ in ATDG. During inference, we set the number of sampling steps to 67 for ATDG and 100 for pure CFG, ensuring the same number of function evaluations (NFE).

\subsubsection{Evaluation Metrics.} We evaluate our models and baselines on audio quality and video-audio alignment, both semantically and temporally. Objective metrics include commonly used Fr\'echet Audio Distance (FAD), Kullback-Leibler (KL) divergence, Inception Score (IS), Fr\'echet Distance (FD) (using the PANN tagging model), Imagebind Score (IBS) \cite{girdhar2023imagebind} and temporal alignment accuracy (AA) employed in Diff-Foley. For subjective evaluation, we recruit 15
human raters to score three aspects: (i) overall audio quality (OVL), (ii) semantic relevance to the input video (S-REL) and (iii) temporal synchronization with the video (T-REL). All scores are rated on a 1–5 scale. More details of these metrics are introduced in Appendix~\ref{app:evaluation}.

\begin{figure*}[h]
  \centering
  \includegraphics[width=1.0\linewidth]{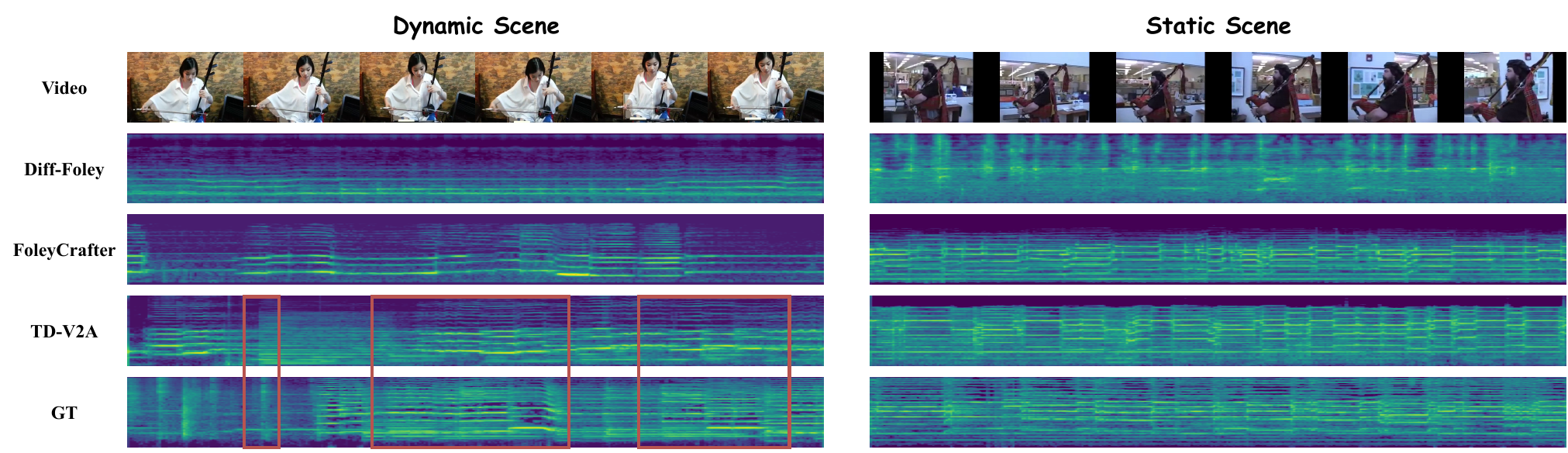}
\caption{Performance comparison across dynamic and static scenes. In the dynamic instrument scene, TD-V2A achieves precise rhythmic synchronization with GT. For the static scene, TD-V2A maintains higher audio quality and semantic fidelity compared to other baselines. We put more samples in Appendix~\ref{app:results}.}
\label{fig:casestudy}
\end{figure*}

\subsection{Main Results}
\label{sec:main_results}
We compare our results with GT (\textit{i.e.}, ground truth) and a variety of V2A systems\footnote{Some recent methods, such as ThinkSound, utilize extra text information (VT2A) and are not directly comparable to standard V2A generation. We further compare with them in Appendix~\ref{app:vt2acomparison}.}, including IM2WAV \cite{sheffer2023hear}, Diff-Foley \cite{luo2023diff}, FoleyGen \cite{mei2024foleygen}, VTA-LDM \cite{xu2024vtaldm}, FoleyCrafter \cite{zhang2024foleycrafter}, Frieren \cite{wang2024frieren}, V2A-Mapper \cite{wang2024v2amapper}, VAB-Encodec \cite{su2024vab}, VATT \cite{liu2024VATT}, MMAudio \cite{cheng2025mmaudio} and AudioX \cite{tian2025audiox}. Detailed descriptions of these methods are provided in Appendix~\ref{app:baseline}. 
The quantitative results are summarized in Table \ref{tab:v2aobjective}. Our method considering HCL, FTD and ATDG outperforms the baseline models, with especially remarkable improvements in the FAD, IS, and IBS metrics. 
More importantly, apart from Diff-Foley, which pretrains contrastive models with AA objective on VGGSound and AudioSet datasets to directly optimize this metric, our method surpasses all other baselines on AA and achieves performance comparable to Diff-Foley. 
In the following ablation studies, we also observe that compared with using condition signals from raw frames, both FTD and CTD achieve better generation quality, further validating the effectiveness of our methods.

\begin{table}[t]
\centering
  \small
  \setlength{\tabcolsep}{3pt}
  \begin{tabular}{c c | c c c c c c}
\toprule
\textbf{Pret.} & \textbf{Condition} & \textbf{FAD $\downarrow$} & \textbf{KL $\downarrow$} & \textbf{IS $\uparrow$} & \textbf{FD $\downarrow$} & \textbf{IBS $\uparrow$} & \textbf{AA $\uparrow$} \\
\midrule
T2A & CLIP & 0.64 & 2.20 & 16.6 & 4.33 & 31.8 & 87.2 \\
T2A & CLIP+CAVP & 0.57 & \textbf{2.16} & 16.4 & 4.06 & 31.9 & 88.5 \\
\midrule
T2A & CLIP+CTD & 0.57 & 2.20 & \textbf{17.1} & 3.93 & 33.3 & 88.3 \\
T2A & CLIP+FTD & 0.55 & 2.18 & 16.5 & 3.98 & 33.4 & 88.8 \\
HCL & CLIP+FTD & \textbf{0.53} & \textbf{2.16} & 16.9 & \textbf{3.79} & \textbf{33.8} & \textbf{89.1} \\
\bottomrule
\end{tabular}
  \caption{Ablation studies on pre-training (denoted as Pret.) and condition.}
\label{tab:ablation_1}
\end{table}

\subsection{Ablation Studies}
\label{sec:ablation}
\subsubsection{Comparison with Different Conditions.} 
To prove the effectiveness of our proposed auxiliary information for V2A, FTD and CTD, we compare their results with our baseline model with pure CLIP and CLIP+CAVP  in Table~\ref{tab:ablation_1}. 
Our methods yield consistent improvements across all evaluation metrics over using only CLIP, with FTD exhibiting more substantial gains. Moreover, the FTD model even achieves better results than CLIP+CAVP, which requires extra training objectives and stages. These results further validate that incorporating TD information, through capturing visual dynamics, enhances the performance of V2A generation.

\subsubsection{Effectiveness of HCL Strategy.}Next, we demonstrate the effectiveness of our HCL strategy by comparing it with pure T2A pre-training. As shown in Table~\ref{tab:ablation_1}, T2A pre-training first provides a strong initialization, allowing the model to learn fundamental audio generation capabilities before incorporating video conditions. Building upon this foundation, our proposed HCL method progressively introduces TD features, enabling the model to develop more fine-grained capabilities while retaining the knowledge acquired in earlier stages. Consequently, HCL achieves further improvements over the already strong T2A-pretrained model.

\begin{table}[t]
\centering
\small
\setlength{\tabcolsep}{3pt}
\begin{tabular}{l | c c c c c c}
\toprule
\textbf{Setting} & \textbf{FAD $\downarrow$} & \textbf{KL $\downarrow$} & \textbf{IS $\uparrow$} & \textbf{FD $\downarrow$} & \textbf{IBS $\uparrow$} & \textbf{AA $\uparrow$} \\
\midrule
Low static ($w_{\text{TD}}=0.5$)  & 0.57 & 2.18 & 16.7 & 3.85 & 33.6 & 88.2 \\
High static ($w_{\text{TD}}=1.5$) & 0.62 & 2.21 & 16.5 & 4.01 & 33.0 & \textbf{89.1} \\
ATDG (Ours) & \textbf{0.53} & \textbf{2.16} & \textbf{16.9} & \textbf{3.79} & \textbf{33.8} & \textbf{89.1} \\
\bottomrule
\end{tabular}
\caption{Ablation of TD guidance scale $w_{\text{TD}}$.}
\label{tab:ablation_w2}
\end{table}

\begin{table}
\centering
  \small
  \begin{tabular}{c | c c c c c c}
\toprule
$k$ & \textbf{FAD $\downarrow$} & \textbf{KL $\downarrow$} & \textbf{IS $\uparrow$} & \textbf{FD $\downarrow$} & \textbf{IBS $\uparrow$} & \textbf{AA $\uparrow$} \\
\midrule
1 & 0.58 & 2.21 & \textbf{17.2} & 4.11 & 33.4 & 88.6 \\
2 & \textbf{0.53} & \textbf{2.16} & 16.9 & \textbf{3.79} & \textbf{33.8} & \textbf{89.1} \\
3 & 0.59 & 2.17 & 16.9 & 3.94 & \textbf{33.8} & 88.8 \\
\bottomrule
\end{tabular}
\caption{Ablation studies on window length $k$.}
\label{tab:ablation_2}
\end{table}

\subsubsection{Impact of ATDG.}We further compare our proposed ATDG with two static $w_{\text{TD}}$ settings in Table~\ref{tab:ablation_w2}. As shown, a small constant $w_{\text{TD}}$ preserves the perceptual quality of the generated audio, but fails to provide sufficient visual difference conditioning, leading to worse temporal synchronization. In contrast, a large static $w_{\text{TD}}$ introduces noticeable distortion and degrades audio quality. Our method, ATDG, effectively achieves a balance by emphasizing TD in the early stages and gradually reducing it in later steps. These results highlight the importance of dynamically annealing the guidance strength for effectively exploiting TD while maintaining high audio quality throughout the diffusion process. We also ablate the frame-level guidance scale $w_f$ in Appendix~\ref{app:guidanceablation}.

\subsubsection{Window Length.} We also investigate the effect of our hyperparameter, namely the window length $k$, in the setting of CLIP+FTD. Specifically, we set $k$ to 1, 2, and 3, and report the results in Table~\ref{tab:ablation_2}. We observe that when $k=1$, the window is too short to effectively capture visual dynamics, while when $k=3$, the window becomes too large and may be sensitive to distracting information in the video. In contrast, $k=2$ serves as a balanced choice, as it not only captures short-term dynamics but also preserves fine-grained details in the video, thereby achieving the best performance.



\section{Conclusion}
In this work, we identify TD as the fundamental cue that distinguishes V2A from I2A generation. Building upon this insight, we show that explicitly exploiting TD provides a simple yet effective paradigm for V2A, enabling strong generation results without additional supervision, strong inductive biases, or substantial architectural modifications. 
Together with HCL and ATDG, TD consistently improves generation quality and even surpasses the additionally developed CAVP model, demonstrating the remarkable potential of explicit TD modeling for future V2A research.

{
\small
\bibliography{aaai2027}
}

\newpage
\clearpage
\appendix
\setcounter{secnumdepth}{1}

\section{Audio Latent Diffusion Model}
\label{app:preliminary}
Diffusion models have advanced cross-modal audio generation with the strong capability to faithfully capture the target distribution~\cite{ho2020denoising,sgms}. 
They are composed of two processes: a~\textit{forward} process transforming the data distribution into a known prior distribution,~\textit{e.g.}, standard Gaussian distribution, and a~\textit{reverse} process gradually reconstructing the target distribution from the prior distribution with iterative sampling steps.
In recent works, audio generation systems are usually built upon a latent diffusion framework.
At early stage, the small space is compressed from the mel-spectrogram and the entire generation process includes latent diffusion sampling, VAE decoding, and a vocoder transforming mel-spectrogram to waveform~\cite{liu2023audioldm,liu2024audioldm,wang2024tiva}.

In this work, we directly develop the diffusion model in the latent space compressed from audio waveform, avoiding cascaded VAE and vocoder.
The details of VAE is introduced in Appendix~\ref{app:vae}.
Given an audio sample $\bm{x}_0\in \mathbb{R}^{L}$ sampled from the observed dataset, where $L$ is the sample length, audio generation systems usually develop a compression network to encode it into a latent representation $\bm{z}_0 \in \mathbb{R}^{T_d \times F_d}$, where $T_d$ and $F_d$ denote the temporal and feature dimension, respectively. 
Then, in the compressed latent space, a forward process of the latent diffusion model injects Gaussian noise $\bm{\epsilon} \sim \mathcal{N}(\mathbf{0}, \mathbf{I})$ into $\bm{z}_0$ with a predefined noise schedule $\beta_t$, constructing noisy representations $\bm{z}_t\in \mathbb{R}^{T_d \times F_d}$ at each time step $t\in (0,T]$ with:
\begin{equation}
q(\bm{z}_t | \bm{z}_0) = \mathcal{N} (\bm{z}_t ; \sqrt{\bar{\alpha}_t} \, \bm{z}_0, \, (1-\bar{\alpha}_t) \mathbf{I}),
\end{equation}
where $\bar{\alpha}_t = \prod_{s=1}^{t} (1-\beta_s)$ indicates the noise level at time step $t$. 
When $t=T$, the clean latent representation $\bm{z}_0$ at time step $t=0$ will be transformed into uninformative Gaussian noise $\bm{z}_T$. 

In generation, the reverse process starts from Gaussian noise $\bm{z}_T$, gradually reconstructing the target $\bm{z}_0$ with an iterative denoising and refinement trajectory:
\begin{equation}
\label{reverse}
p_\theta(\bm{z}_{t-1} | \bm{z}_t) = \mathcal{N} \big(\bm{z}_{t-1} ; \bm{\mu}_\theta(\bm{z}_t, t), \, \sigma_t \mathbf{I}\big),
\end{equation}
where the time-dependent variation $\sigma_t$ is usually predefined according to $\beta_t$ and the mean function $\bm{\mu}_\theta(\bm{z}_t, t)$ can be parameterized with different methods and predicted by a neural network.
For diffusion-based audio generation systems~\cite{liu2023audioldm,evans2025stable}, noise prediction is a popular parameterization method, where the network is optimized with the objective:
\begin{equation}
\mathcal{L}_{\text{audio}} = \mathbb{E}_{\bm{z}_0, \bm{\epsilon}, t} \Big[ \big\| \bm{\epsilon} - \bm{\epsilon}_\theta(\sqrt{\bar{\alpha}_t} \, \bm{z}_0 + \sqrt{1-\bar{\alpha}_t} \, \bm{\epsilon}, t) \big\|_2^2 \Big].
\end{equation}
Given enough sampling steps, the target representation $\hat{\bm{z}}_0$ can be generated from $\bm{z}_T$, and then the audio waveform $\hat{\bm{x}}_0$ can be decoded from $\hat{\bm{z}}_0$ with the compression network.

\section{Impact of Frame-Level Guidance Scale}
\label{app:guidanceablation}

\begin{figure}[t]
  \centering
  \includegraphics[width=1.0\linewidth]{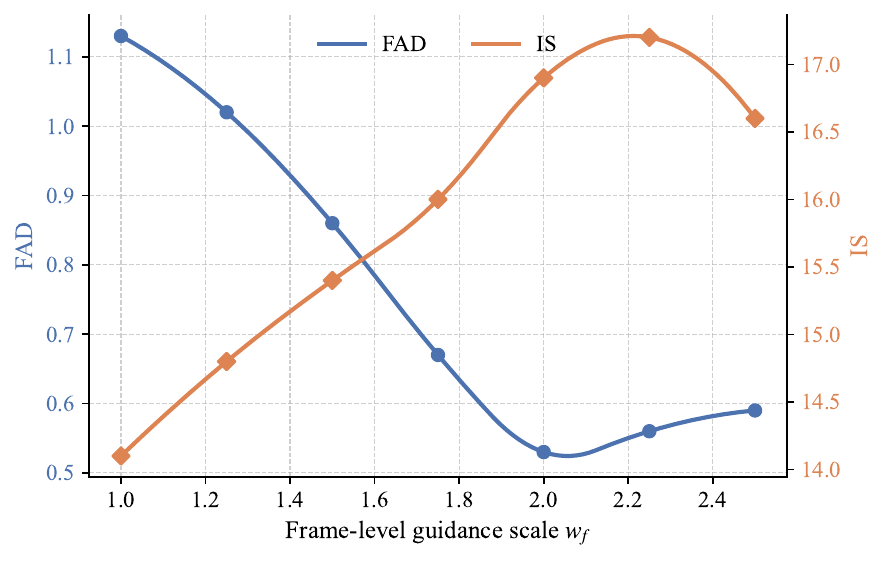}
\caption{Ablation study on frame-level guidance scale $w_f$.}
\label{fig:ablation_wf}
\end{figure}

Besides guidance scale of TD $w_{\text{TD}}$, we also analyze the impact of frame-level guidance scale $w_f$. Specifically, we vary its value from 1.0 to 2.5, with results shown in Figure~\ref{fig:ablation_wf}. In general, increasing $w_f$ within a moderate range improves audio quality; however, exceeding the optimal value of 2.0 leads to a decline in audio fidelity.

\section{Datasets}
\label{app:datasets}
We present the datasets used to pre-train and fine-tune our model in Table~\ref{tab:app-datasets}. AudioCaps~\cite{kim2019audiocaps} serves as a widely used benchmark for audio captioning that comprises 44K audio segments sourced from AudioSet, each annotated with human-written textual descriptions. AudioSet~\cite{gemmeke2017audio} is a large-scale, weakly labeled dataset released by Google, containing more than 2M 10-second audio clips across over 600 sound event categories. VGGSound~\cite{chen2020vggsound} is another large-scale dataset, consisting of over 200K 10-second video clips collected from YouTube and covering 310 distinct sound categories. FreeSound~\cite{font2013freesound} is a collaborative online platform that provides a diverse collection of user-uploaded audio samples, commonly utilized for environmental sound classification and retrieval tasks. The Million Song Dataset (MSD)~\cite{bertin2011million} provides metadata along with pre-extracted audio features for 1M popular music tracks, facilitating large-scale music analysis and recommendation research.

\begin{table}[h]
\centering
\small
\setlength{\tabcolsep}{3pt}
\begin{tabular}{c | c c c}
\toprule
\textbf{Dataset} & \textbf{Task} & \textbf{Hours (h)} & \textbf{Source} \\
\midrule
AudioCaps  & T2A & 109 & \cite{kim2019audiocaps} \\
AudioSet & T2A, V2A & 5800 & \cite{gemmeke2017audio} \\
VGGSound & T2A, V2A & 550 & \cite{chen2020vggsound}  \\
FreeSound & T2A & 6246 & \cite{font2013freesound} \\
MSD & T2A & 7333 & \cite{bertin2011million} \\
\bottomrule
\end{tabular}
\caption{Details of training datasets in TD-V2A.}
\label{tab:app-datasets}
\end{table}

\section{Compression Network}
\label{app:vae}
We adopt a waveform-domain VAE based on the Oobleck framework with a sampling rate of 16 kHz. 
Previous V2A generation approaches may rely on Mel-spectrograms as intermediate representations, typically following the AudioLDM~\cite{liu2023audioldm} paradigm where diffusion models are developed in the continuous latent space of the Mel-spectrogram and a separate vocoder is required to reconstruct waveforms. 
In contrast, we directly compress audio waveforms into continuous latent representations, avoiding the cascaded VAE and vocoder pipeline. 
To ensure fair comparison with existing audio generation systems, the waveform compression architecture and training procedure follow the widely adopted and publicly available Stable Audio Open~\cite{evans2025stable} framework.

The encoder and decoder are designed symmetrically with a base channel size of 128 and channel multipliers of 1, 2, 4, 8, and 16. The corresponding stride factors are 2, 2, 4, 4, and 10, resulting in an overall temporal down-sampling ratio of 640. 
The encoder maps the input mono-channel waveform into a 128-dimensional latent representation, while the decoder reconstructs the waveform from a 64-dimensional latent code through a variational bottleneck layer. Snake activation functions are applied throughout the network, and no final tanh activation is used in the decoder.

\section{Model Configurations}
\label{app:ldm}
Our diffusion framework is based on the Diffusion Transformer (DiT) architecture and follows the latent diffusion modeling (LDM) paradigm, which provides strong generative performance along with effective context representation. During T2A pre-training, we employ FLAN-T5~\cite{chung2024scaling} as the text encoder, and adopt the VAE in Appendix~\ref{app:vae} to compress the original waveform into a compact latent space. The diffusion backbone utilizes a DiT configuration with 24 layers and 24 attention heads, each with an embedding dimension of 1536. The model incorporates both cross-attention and global conditioning mechanisms: cross-attention is used for all conditional inputs, while global conditioning is dedicated to handling duration-related control signals. The internal token dimension of the diffusion model is set to 64, with a conditional token dimension of 768 and a global conditioning embedding dimension of 1536. The output latent representation shares the same dimensionality as io\_channels, which is 64.

\section{Baseline Methods}
\label{app:baseline}
We introduce the baseline methods used for comparison in Table~\ref{tab:v2aobjective}.

\textbf{IM2WAV}~\cite{sheffer2023hear} is an image-guided open-domain audio generation model that employs two transformer language models over discrete audio tokens derived from a VQ-VAE, with CLIP-based visual conditioning and classifier-free guidance.

\textbf{Diff-Foley}~\cite{luo2023diff} is a V2A synthesis approach based on latent diffusion models, leveraging contrastive audio-visual pretraining (CAVP) and cross-attention with double guidance to achieve strong temporal and semantic alignment.

\textbf{FoleyGen}~\cite{mei2024foleygen} formulates V2A generation as a language modeling task, using a neural audio codec and a single transformer conditioned on visual features to generate audio tokens. It further introduces novel visual attention mechanisms to improve temporal alignment between generated audio and video content.


\textbf{VTA-LDM}~\cite{xu2024vtaldm} is a latent diffusion-based V2A framework that investigates key factors such as vision encoders, auxiliary embeddings, and data augmentation. It achieves strong performance in generating semantically and temporally aligned audio through systematic design and evaluation.

\textbf{FoleyCrafter}~\cite{zhang2024foleycrafter} is a V2A framework built upon pretrained T2A models, introducing a semantic adapter and a temporal controller to enhance semantic relevance and precise audio-video synchronization, while supporting controllable generation via text prompts.

\textbf{V2A-Mapper}~\cite{wang2024v2amapper} is a lightweight approach that bridges visual and audio modalities by mapping CLIP embeddings to CLAP space. Conditioned on the translated embeddings, a pretrained AudioLDM generates high-fidelity and well- aligned audio with minimal training cost.


\textbf{VAB-Encodec}~\cite{su2024vab} is a unified audio-visual framework that performs both representation learning and generation in latent space. It leverages visual-conditioned masked audio token prediction and iterative decoding to produce high-quality, semantically aligned audio while supporting various downstream tasks.

\textbf{VATT}~\cite{liu2024VATT} is a controllable V2A framework that generates audio from video with optional text guidance and can produce audio captions. It uses a fine-tuned LLM to map video features and a bidirectional transformer to generate audio tokens, which are decoded into waveforms, enabling text-guided audio generation and captioning.

\textbf{MMAudio}~\cite{cheng2025mmaudio} is a V2A framework that adopts a multimodal joint training paradigm over audio-video and audio-text datasets. By jointly learning from video, audio, and text with modality masking and a conditional synchronization module, it generates high-quality audio with strong semantic alignment and temporal synchronization.

\textbf{AudioX}~\cite{tian2025audiox} is a unified multimodal audio generation framework that supports flexible conditioning on text, video, and audio inputs. It employs a Multimodal Adaptive Fusion module and is trained on a large-scale dataset to achieve strong cross-modal alignment and high-quality generation across diverse tasks.

For Diff-Foley, VTA-LDM and FoleyCrafter, we generate audio samples using their official implementations. For V2A-Mapper, it supplies pre-generated audio samples for evaluation. For FoleyGen, VAB-Encodec, VATT and AudioX, we report the results presented in their original papers. The IM2WAV and MMAudio results are adopted from VATT and Omni2Sound, respectively.

\begin{figure*}[t]
  \centering
  \includegraphics[width=1.0\linewidth]{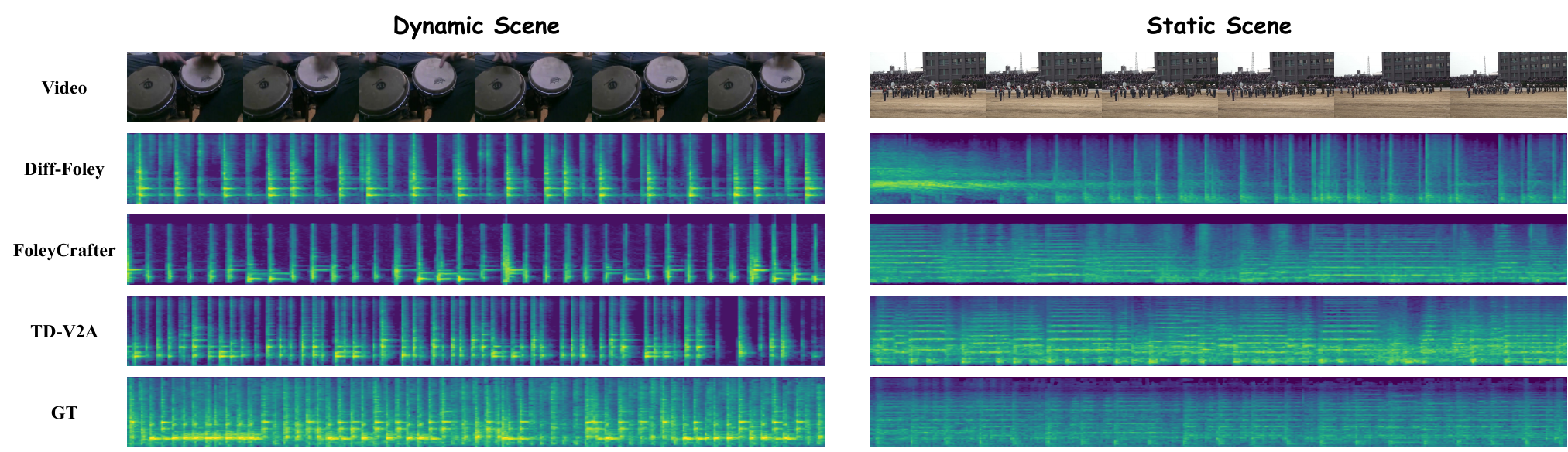}
\caption{Performance comparison across dynamic and static scenes.}
\label{fig:app_case}
\end{figure*}

\begin{figure}[t]
  \centering
  \includegraphics[width=1.0\linewidth]{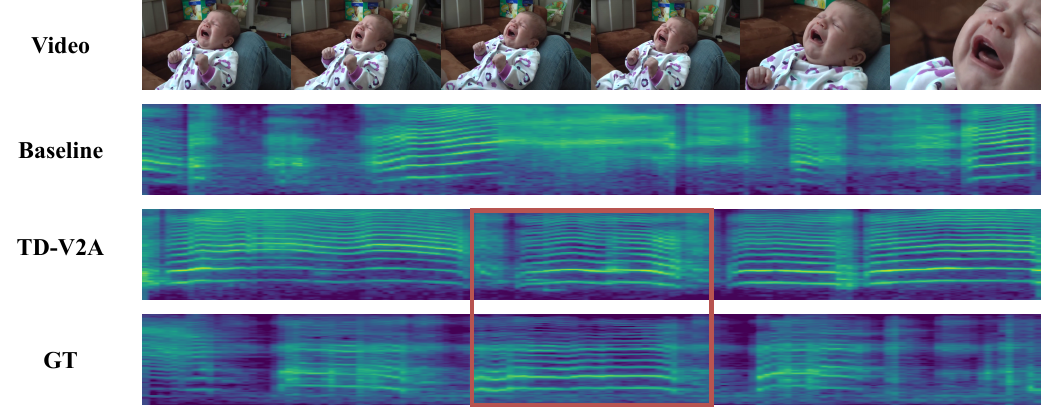}
\caption{Spectrogram comparison for the "baby crying" video. TD-V2A achieves significantly better temporal synchronization and semantic alignment with the GT than the T2A+CLIP baseline.}
\label{fig:inferencemethodcase}
\end{figure}

\begin{table}[t]
\centering
  \small
  \setlength{\tabcolsep}{4pt}
  \begin{tabular}{l | c c c c c}
\toprule
\textbf{Method} & \textbf{FAD $\downarrow$} & \textbf{KL $\downarrow$} & \textbf{IS $\uparrow$} & \textbf{FD $\downarrow$} & \textbf{IBS $\uparrow$} \\
\midrule
ThinkSound (w/o CoT) & 1.10 & 1.60 & 11.7 & 7.41 & 26.0 \\
HunyuanVideo-Foley & 2.36 & 1.74 & 11.6 & 10.02 & 32.0 \\
AudioX (VT2A) & 1.24 & \underline{1.59} & 14.9 & 8.29 & 26.0 \\
MMAudio (w/ text) & \underline{0.91} & 1.63 & 13.4 & 5.28 & 29.0 \\
Omni2Sound (VT2A) & \textbf{0.53} & \textbf{1.35} & \underline{15.8} & \textbf{2.95} & \textbf{34.0} \\
\midrule
TD-V2A (Ours) & \textbf{0.53} & 2.16 & \textbf{16.9} & \underline{3.79} & \underline{33.8} \\
\bottomrule
\end{tabular}
  \caption{Comparison of our method with VT2A methods in Omni2Sound. Note that these methods utilize the carefully curated text caption in VGGSound-Omni.}
\label{tab:vt2acomparison}
\end{table}

\section{Evaluation Metrics}
\label{app:evaluation}

\subsection{Objective Metrics}
We introduce the objective metrics employed in our evaluation, including Fr\'echet Audio Distance (FAD), Kullback-Leibler (KL) divergence, Inception Score (IS), Fr\'echet Distance (FD), Imagebind Score (IBS)~\cite{girdhar2023imagebind} and temporal alignment accuracy (AA)~\cite{luo2023diff}. 

FAD, adapted from FID (Fr\'echet Inception Distance), measures the distributional gap between generated and reference audio using VGGish embeddings, serving as the main measure of audio fidelity. KL divergence evaluates the difference between acoustic event posteriors of the generated and ground truth audio. IS captures both diversity and specificity of generated samples by computing entropy over class predictions. FD, similar in formulation to FAD, is calculated in more general embedding spaces to provide an additional assessment of generative quality.

ImageBind Score (IBS) evaluates the semantic alignment between generated audio and the corresponding video by computing the cosine similarity of their embeddings in a shared multimodal space, using the ImageBind model. A higher IBS indicates stronger semantic correlation between audio and visual content.

Temporal Alignment Accuracy (AA) measures how well generated audio is synchronized with video by classifying correctly aligned versus temporally shifted or mismatched audio-video pairs. The percentage of true pairs correctly identified serves as a quantitative measure of temporal synchronization.

\subsection{Subjective Metrics}
We randomly select 20 samples from the VGGSound test set for the subjective evaluation. Each sample group contains results from Diff-Foley, FoleyCrafter, and the ground truth (GT), with the order of items within each group randomly shuffled. Every group is evaluated by 15 human raters. Participants rate three aspects—overall audio quality (OVL), semantic relevance to the input video (S-REL), and temporal synchronization with the video (T-REL)—on a 1-to-5 scale. For OVL, raters assess perceptual audio quality, while S-REL and T-REL reflect the semantic alignment and temporal correspondence of the audio to the video, respectively. All ratings were given in integer increments.

\section{Comparison with VT2A methods}
\label{app:vt2acomparison}
Some recent methods, such as HunyuanVideo-Foley~\cite{shan2025hunyuanvideofoley}, ThinkSound~\cite{thinksound}, and Omni2Sound~\cite{omni2sound}, incorporate additional text conditions during inference, and therefore considered as video-and-text-to-audio (VT2A) generation rather than standard V2A. In this section, we further compare our method with several representative VT2A approaches reported in Omni2Sound. As shown in Table~\ref{tab:vt2acomparison}, despite relying solely on video inputs, our proposed TD-V2A achieves performance comparable to or even surpassing these VT2A methods. Notably, the VGGSound-Omni dataset introduced by Omni2Sound constructs captions through an elaborate caption generation and verification pipeline, whereas our method requires no text annotations at inference. These results further demonstrate the effectiveness of TD-V2A.

\section{More Generation Results}
\label{app:results}
We give more generated cases in Figure~\ref{fig:app_case} and Figure~\ref{fig:inferencemethodcase}.

\end{document}